# *GZSL-MoE:* Apprentissage Généralisé Zéro-Shot basé sur le Mélange d'Experts pour la Segmentation Sémantique de Nuages de Points 3D Appliqué à un Jeu de Données d'Environnement de Collaboration Humain-Robot

Ahed ALBOODY[1], Remi YACOUB[2], DELAPORTE Alexis[3], Lamyae MENHAJ[4]

[1] CESI LINEACT, CESI Nice Campus, France

[2] Université Côte D'AZUR, Polytech Nice-SOPHIA, Nice, France

[3] CESI École d'ingénieurs, CESI Aix-en-Provence Campus, France

[4] CESI École d'ingénieurs, CESI Nice Campus, France

aalboody@cesi.fr, remi.yacoub@etu.univ-cotedazur.fr, alexis.delaporte@viacesi.fr, lmenhaj@cesi.fr

## Résumé

*L'approche d'apprentissage génératif zéro-shot ou zéro exemple de données (Generative Zero-Shot Learning, GZSL) a démontré un potentiel significatif dans les tâches de la segmentation sémantique de nuages de points 3D. GZSL approche exploite des modèles génératifs comme les GAN pour synthétiser des caractéristiques réalistes (caractéristiques réelles) des classes non vues. Cela permet au modèle d'étiqueter des classes non vues lors des tests, bien qu'il ait été entraîné uniquement sur des classes vues. Dans ce contexte, nous présentons le modèle d'apprentissage généralisé zero-shot basé sur le mélange d'experts (GZSL-MoE). Ce modèle intègre des couches de mélange d'experts (MoE) pour générer des caractéristiques artificielles (fake features) qui ressemblent étroitement aux caractéristiques réelles (real features) extraites à l'aide d'un modèle KPConv (Kernel Point Convolution) pré-entraîné sur des classes vues. La principale contribution de cet article est l'intégration du mélange d'experts dans les composants Générateur et Discriminateur du modèle d'apprentissage génératif zéro-shot pour la segmentation sémantique de nuages de points 3D, appliquée au jeu de données COVERED (CollabOratiVE Robot Environment Dataset). COVERED est conçu pour des environnements de collaboration Humain-robot. En combinant le modèle d'apprentissage génératif zéro-shot avec le mélange d'experts, le nouveau modèle GZSL-MoE pour la segmentation sémantique de nuages de points 3D offre une solution prometteuse pour comprendre des environnements 3D complexes, en particulier lorsque des données d'entraînement complètes pour toutes les classes d'objets ne sont pas disponibles. L'évaluation des performances du modèle GZSL-MoE met en évidence sa capacité à améliorer les performances sur les classes vues et non vues.*

## Mots-clés

*Apprentissage Généralisé Zéro-Shot (GZSL), Nuage de points 3D, Segmentation sémantique 3D, Collaboration Humain-robot, COVERED (Jeu de données d'environnement robotique collaboratif), KPConv, Mélange d'experts*

## Abstract

*Generative Zero-Shot Learning approach (GZSL) has demonstrated significant potential in 3D point cloud semantic segmentation tasks. GZSL leverages generative models like GANs or VAEs to synthesize realistic features (real features) of unseen classes. This allows the model to label unseen classes during testing, despite being trained only on seen classes. In this context, we introduce the Generalized Zero-Shot Learning based-upon Mixture-of-Experts (GZSL-MoE) model. This model incorporates Mixture-of-Experts layers (MoE) to generate fake features that closely resemble real features extracted using a pre-trained KPConv (Kernel Point Convolution) model on seen classes. The main contribution of this paper is the integration of Mixture-of-Experts into the Generator and Discriminator components of the Generative Zero-Shot Learning model for 3D point cloud semantic segmentation, applied to the COVERED dataset (CollabOratiVE Robot Environment Dataset) for Human-Robot Collaboration (HRC) environments. By combining the Generative Zero-Shot Learning model with Mixture-of-Experts, GZSL-MoE for 3D point cloud semantic segmentation provides a promising solution for understanding complex 3D environments, especially when comprehensive training data for all object classes is unavailable. The performance evaluation of the GZSL-MoE model highlights its ability to enhance performance on both seen and unseen classes.*

# 1    Introduction

L'apprentissage Zero-Shot (Zero-Shot Learning [1, 2, 3]) permet aux modèles de classifier et de segmenter des objets appartenant à des classes tant familières qu'inconnues, en s'attaquant au problème des données étiquetées limitées dans les tâches de perception 3D. Alors que l'apprentissage Zero-Shot traditionnel utilise des plongements de mots (word embedding's) ou des attributs sémantiques pour décrire les caractéristiques des objets (caractéristiques artificielles) pour les classes vues et non vues, *l'apprentissage zéro généralisé (Generalized Zero-Shot Learning GZSL)* [3, 4, 5, 6, 7] vise à entraîner un modèle pour classifier des échantillons de données dans des conditions où certaines classes de sortie sont inconnues pendant l'apprentissage supervisé. Pour relever ce défi complexe, le GZSL exploite [1] les informations sémantiques des classes observées (source, vus ou connues) et non observées (cible, non vus ou inconnue) afin de combler l'écart entre ces deux types de classes. Une première approche générative GAN est proposée dans [3] pour l'apprentissage zéro généralisé (3D GZSL) sur des données 3D LiDAR (Light Detection and Ranging) pour la perception 3D, capable de gérer à la fois la classification et, pour la première fois, la segmentation sémantique 3D LiDAR. Les jeux de données-les nuages de points 3D LiDAR contiennent un nombre limité de classes d'objets et de scènes, avec peu de variation intra-classe. Une option pour pallier les limitations des données nuages de points issues des capteurs 3D est de tenter de faire des prédictions au moment de l'inférence pour des objets non vus pendant l'entraînement, en se basant sur des informations auxiliaires concernant les classes non annotées [2], [6]. L'approche générative 3D GZSL pour la segmentation sémantique est appliquée à des scènes extérieures complexes pour la conduite autonome où des objets spécifiques à un pays (véhicules, barrières de chantier, obstacles routiers potentiels...) doivent être largement collectés et étiquetés.

L'application des modèles 3D GZSL sur des données Lidar pour des environnements intérieures complexes, des environnements de collaboration Humain-robot et des environnements industrielles [8], [9], reste une problématique de recherche car les données dynamiques et les actions de collaboration Humain-robot doivent être largement identifiées, collectés et étiquetés.

Les approches génératives GZSL [6], [10] génèrent des caractéristiques artificielles (fake features) de classes non vues avec des caractéristiques sémantiques pour entraîner le classificateur afin de réaliser le ZSL. Cependant, les méthodes génératives nécessitent des efforts d'entraînement supplémentaires lorsque de nouvelles catégories non vues apparaissent, ce qui limite leur capacité de généralisation. De plus, comme les caractéristiques 3D sont plus complexes que les caractéristiques 2D, la distribution des caractéristiques générées ne correspond pas bien à la distribution originale, entraînant des résultats médiocres.

Nos contributions sont les suivantes. (1) nous proposons un cadre génératif général le modèle *GZSL-MoE* pour la segmentation sémantique des nuages de points 3D des environnements intérieurs complexes de collaboration Humain-robot (HRC). (2) Nous appliquons le modèle 3D *GZSL-MoE* sur le benchmark COVERED *(CollabOratiVE Robot Environment Dataset)* pour la segmentation sémantique des données 3D LiDAR.

## 1.1    Formulation du problème de GZSL

Nous supposons que $S = \{(x_{is}, a_{is}, y_{is})\}_{i=1}^{N_s}$ représentant les classes vues (**Seen classes**) et $U = \{(x_{ju}, a_{ju}, y_{ju})\}_{j=1}^{N_u}$ représentant les classes non vues (**Unseen classes**). Les éléments de ces ensembles sont définis comme suit : $x_{is}, x_{ju} \in R^D$ représentent les caractéristiques visuelles des données dans l'espace des caractéristiques X. Ces caractéristiques peuvent être obtenues à l'aide d'un modèle d'apprentissage profond pré-entraîné (dans notre article le modèle KPConv [11] pour la segmentation sémantique des nuages de points 3D). $a_{is}, a_{ju} \in R^K$ représentent les caractéristiques sémantiques (par exemple, les attributs ou les vecteurs de mots) dans l'espace sémantique A. $y_{is} \in Y_s, y_{ju} \in Y_u$ sont les étiquettes des classes vues **S** (**Seen classes**) et non vues **U** (**Unseen classes**) respectivement, où $Y_s$ et $Y_u$ sont les ensembles d'étiquettes pour les classes vues et non vues. Les ensembles d'étiquettes $Y_s$ et $Y_u$ sont disjoints, c'est-à-dire $Y_s \cap Y_u = \emptyset$, et leur union $Y_s \cup Y_u$ représente l'ensemble complet des classes.

*L'objectif du GZSL* (voir Figure 1) est d'apprendre un modèle $f_{\text{GZSL}}$:X→Y capable de classifier des échantillons de test : $D_{test} = \{(x_m, y_m)\}_{m=1}^{N_{test}}$, où $x_m \in R^D$ et $y_m \in Y$

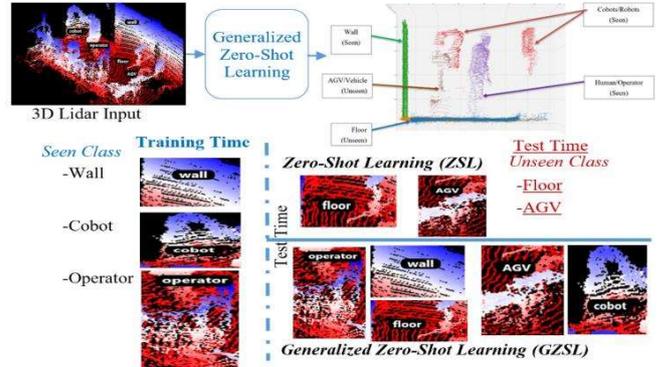

Figure 1. Illustration de la différence entre les deux approches génératives ZSL et GZSL pour des données Lidar de 5 classes (objets) : *Wall, Cobot et Operator (Human)* représentent le classes vues **S** (**Seen classes**) dans la phase d'entrainement, et les deux classes non vues **U** (**Unseen classes**) sont *AGV (an Automated Guided Vehicle)* et *Floor*.

L'apprentissage dans le cadre du GZSL[4] peut être divisé en deux grandes catégories : *l'apprentissage inductif* [4, 6] utilise uniquement les caractéristiques visuelles et les informations sémantiques des classes vues pour construire le modèle. Et en plus des informations des classes vues, *l'apprentissage transductif* [4] utilise également des données non étiquetées des classes non vues pour améliorer l'alignement entre les espaces visuels et sémantiques.

Dans la section suivante, nous allons réaliser un état de l'art sur les méthodes de l'apprentissage GZSL pour la segmentation sémantique des données LiDAR.

## 1.2 Revue de la littérature

***Zero-Shot Learning (GZSL)*** : L'apprentissage zéro-shot (ZSL) [1, 2, 3] peut être considéré comme un cas particulier de l'apprentissage par transfert, où les connaissances issues d'un domaine source (classes vues) et d'une tâche source (classification ou segmentation) sont transférées vers un domaine cible (classes non vues) avec une tâche cible (espace d'étiquettes différent). Nous passons en revue certaines méthodes existantes [1, 2, 3]. Une méthode d'apprentissage zéro-shot multimodal a été proposé dans [10] pour mieux exploiter les informations complémentaires des nuages de points et des images, afin d'obtenir un alignement visuel-sémantique plus précis.

***Generative Zero-Shot Learning (GZSL):*** Pour la segmentation sémantique de nuages de points 3D LiDAR, à notre connaissance, seul l'article [6] propose des solutions en apprentissage génératif/généralisé zéro-shot (GZSL). Le modèle de l'apprentissage génératif zéro-shot (3DGenZ) a émergé comme une approche puissante pour la segmentation sémantique des nuages de points 3D, répondant au défi de classifier et segmenter des objets appartenant à des classes inconnues. 3DGenZ combine les avantages de l'apprentissage zéro-shot avec des modèles génératifs afin d'améliorer les performances sur les classes vues et non vues.

L'architecture GZSL [6] repose sur trois blocs principaux, entraînés séquentiellement comme suit : un extracteur de caractéristiques de type Backbone pour traiter les nuages de points 3D, un bloc génératif entrainé pour générer des caractéristiques conditionnées par un prototype de classe, puis un classificateur qui prédit une classe à partir d'une caractéristique inférée. L'architecture GZSL [6] suit une procédure d'inférence en quatre étapes d'apprentissage: (1) *Feature Extraction Backbone* bloc pour l'extraction des caractéristiques sur les classes vues *S* (*Seen classes*) où le modèle de base est entraîné sur les classes vues pour apprendre les caractéristiques fondamentales ; (2) *Générateur bloc* où le générateur est entraîné à produire des caractéristiques artificielles pour les classes non vues *U* (*Unseen classes*) ; (3) *Classificateur bloc* où le classificateur est d'abord entraîné en utilisant les caractéristiques artificielles des classes non vues avec l'apprentissage zéro-shot (ZSL), puis il est entraîné sur les classes vues *S* et non vues *U* avec l'apprentissage généralisé zéro-shot (GZSL), et (4) *L'inférence* se produit à partir du modèle de base jusqu'au classificateur final, permettant la reconnaissance des classes vues *S* et non vues *U*. *3DGenZ génère* toutes des caractéristiques artificielles des classes non vues en utilisant des caractéristiques sémantiques pour entraîner le classificateur et permettre le transfert zéro-shot. Cependant, les méthodes génératives nécessitent des efforts d'entraînement supplémentaires lorsque de nouvelles classes non vues apparaissent, ce qui limite leur capacité de généralisation.

***Mixture-of-Experts (MoE)***: Les couches de Mélange d'Experts (MoE) dans Mistral [12], [13], [14], DeepSeek [14], [15] sont utilisées dans des modèles génératifs pour des applications du traitement de langage naturel. Récemment, des nouvelles architectures MoE sont proposées dans [16], [17] pour la reconnaissance de gestes des mains. Dans la couche MoE [12], [16], [17], chaque vecteur d'entrée est strictement assigné à K experts parmi M experts en total par un routeur. Les deux experts sélectionnés (par la porte-the Gate) auront leur somme pondérée [12], [16], [17] comme sortie de la couche pour ces mêmes deux experts. En dessous, un expert est défini par un bloc standard de *FeedForward Neural (FFN)*. Une collection de sous-modèles (groupe d'experts utilisant des couches MoE spécialisées) optimisera leurs performances, et ce, avec un coût computationnel constant [12], [16], [17]. Les architectures Mixture-of-Experts (MoE) [1, 3, 6–9] gagnent en popularité dans le domaine des grands modèles de langage (LLM) et de la vision par ordinateur. le modèle Mistral 7B MoE de Mistral [12], [13], [14] utilise une architecture MoE avec 8 experts, dont seulement 2 sont activés pour chaque entrée, permettant ainsi de bénéficier d'un grand modèle tout en gardant un coût d'inférence réduit. DeepSeek MoE [14], [15] expérimente avec différentes configurations d'experts et de routeurs, combinant des approches pour maximiser l'efficacité énergétique et la performance sur des tâches complexes de génération de texte.

***Ces architectures MoE*** ne sont pas explorées pour traiter de données de nuages de points 3D LiDAR de haute dimension pour la segmentation sémantique et la classification en robotique dans des environnements intérieurs complexes. Donc, nous proposerons d'intégrer des architectures MoE de Mistral [12], [13], [14] et nos travaux dans avec des modèles GZSL basés sur les GAN pour traiter efficacement la segmentation sémantique et la classification de données 3D LiDAR de haute dimension. L'introduction des couches MoE dans le modèle génératif GZSL (dans le générateur et le discriminateur à la fois) génère des caractéristiques artificielles des classes non vues en utilisant des caractéristiques sémantiques des séquences complexes de nuages de points 3D. Cette technique permet une gestion évolutive et efficace pour améliorer la capacité de généralisation du modèle génératif GZSL et pour mieux entraîner le discriminateur ci qui va permettre de séparer entre les classes vues et non vues, et va être capable de classifier des échantillons de test lors de la phase de test.

A notre connaissance l'apprentissage GZSL pour la segmentation sémantique des données LiDAR n'est pas encore utilisée et explorée en robotique pour des environnements intérieurs complexes de collaboration Humain-robot. Donc, nous allons appliquer le modèle proposé *3D GZSL-MoE sur le benchmark COVERED (CollabOratiVE Robot Environment Dataset)* pour la segmentation sémantique des données 3D LiDAR.

## 2 Contribution : 3D GZSL-MoE

Dans cette section, nous allons proposer le modèle GZSL-MoE en tant que modèle d'apprentissage généralisé zéro-shot basé sur le mélange d'experts pour la segmentation sémantique de nuages de points 3D appliqué à un jeu de données d'environnement de collaboration Humain-robot.

Nous nous inspirons de l'architecture 3D GZSL [6] pour proposer un modèle **3D GZSL-MoE** (voir Figure 2) en quatre étapes d'apprentissage: (1) *Feature Extraction Backbone bloc* pour l'extraction des caractéristiques réelles (**Real Feature**) sur les classes vues $S$ (**Seen classes**) où le modèle de base est entraîné sur les classes vues $S$ pour identifier les classes par *le MoE Classificateur1 bloc*; (2) *GMoE Générateur bloc (GMoE Feature Generator)* où le générateur est **un modèle MoE** qui est entraîné à produire des caractéristiques artificielles (**Fake Feature**) pour les classes vues $S$ (**Seen classes**) et non vues (**Unseen classes**); (3) *DMoE Discriminateur (classificateur 2) bloc* est d'abord entraîné sur des classes non vues avec l'apprentissage zéro-shot (ZSL), puis il est entraîné sur les classes vues $S$ et non vues $U$ avec *l'apprentissage généralisé zéro-shot (GZSL)*, et(4) *L'inférence* se produit à partir du modèle de base jusqu'au *DMoE discriminateur (classificateur 2)*, permettant la reconnaissance des classes vues $S$ et non vues $U$. La contribution consiste à introduire des MoE couches dans le modèle d'apprentissage GZSL qui combine les connaissances de plusieurs modèles FFN spécialisés, appelés "experts", des couches MoE, pour améliorer la performance générale au sein de GZSL. Dans l'architecture GZSL-MoE, MoE est composé de plusieurs réseaux "experts", et un mécanisme de porte détermine si un expert donné est activé pour une entrée donnée LiDAR. Cela permet au modèle GZSL d'améliorer leur capacité de généralisation et de prendre en compte des données plus grandes et plus complexes de manière plus efficace, en utilisant de manière optimale les connaissances spécialisées des différents experts.

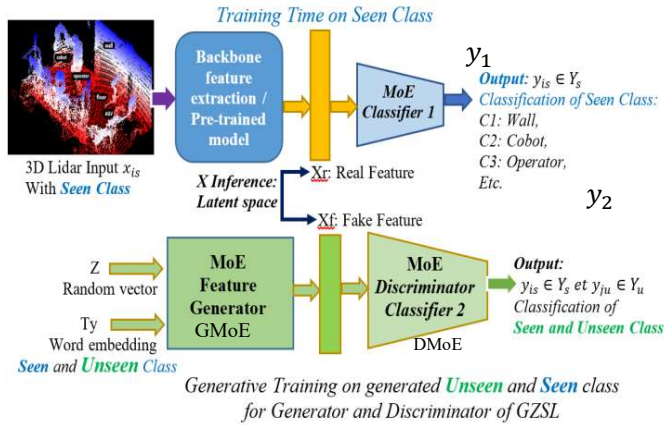

Figure 2. Architecture 3D GZSL-MOE : modèle d'apprentissage généralisé zéro-shot (GZSL) basé sur le mélange d'experts (MoE).

## 2.1 Description de l'architecture GZSL-MoE

Tout d'abord, en entrée, nous avons des données de nuages de points 3D (***Point Cloud Data, PCD***). Soit $C$ un ensemble de classes des nuages de points représenté par $P$ des nuages de points 3D, $P = (P_i)_{i \in I}$ un ensemble de classes des nuages de points 3D, et $Y = (Y_i)_{i \in I}$ l'ensemble des étiquettes de classes correspondantes pour la segmentation sémantique.

L'ensemble des classes d'objets $C$ est divisé en classes vues $S$ et classes non vues $U$. Les points 3D appartenant aux classes vues $S$ sont notés $P^S = (P_i)_{i \in I^S}$ avec $I^S = \{i \in I | y_i \in S\}$ et leurs étiquettes de classe correspondantes $Y^S = (Y_i)_{i \in I^S}$ ; et de même pour les classes non vues $U : P^U = (P_i)_{i \in I^U}$ avec $I^U = \{i \in I | y_i \in S\}$ leurs étiquettes de classe correspondantes $Y^U = (Y_i)_{i \in I^U}$ . Pendant l'entraînement, seuls les points 3D appartenant aux classes vues $S$ ($P^S = (P_i)_{i \in I^S}$ et leurs étiquettes de classe correspondantes $Y^S = (Y_i)_{i \in I^S}$ sont disponibles en entrée pour entrainer *un modèle Backbone Feature comme KPconv* [11], [19]. En entrée de *MoE Générateur bloc,* nous utilisons l'embedding de mots (Word Embedding), les prototypes de classe pour entrainer à partir de $P^S$ et généraliser à $P^U$ sans voir aucun exemple d'une classe de $U$ est impossible sans connaissance supplémentaire. Pour cela, nous devons nous appuyer sur des informations auxiliaires : les *prototypes de classe*. Ces prototypes ne sont pas des exemples concrets (puisqu'aucun *shot* n'est autorisé), mais des vecteurs d'embedding de dimension $D$. Ils sont notés : $T = T_y = \{t_c \in R^D | c \in C\}$ où chaque classe possède un unique prototype. Nous distinguons $T^S$ et $T^U$, qui sont les sous-ensembles de T correspondant respectivement aux classes vues $S$ et aux classes non vues $U$. Ces prototypes servent de représentations vectorielles des classes et permettent d'établir des correspondances entre les classes vues ($T^S$) et les classes non vues ($T^U$). Grâce aux word embeddings (comme Word2Vec ou GloVe), chaque classe est associée à un vecteur d'embedding $t_c \in R^D$. Ces prototypes de classe fournissent une information sémantique qui aide à la généralisation vers les classes non vues en exploitant des relations entre concepts similaires.

Pour étudier la représentation des classes de nuages de points 3D. L'objectif est d'encoder des classes de nuages de points $P = (P_i)_{i \in I}$ et les prototypes de classe de $T = T_y$ dans un espace de représentation commun X (un espace latent pour l'inférence), où les classes de nuages de points 3D et les prototypes appartenant à une même classe possèdent des embeddings similaires. L'embedding des classes P de nuages de points 3D est défini par le bloc *Feature Extraction Backbone* généralement implémentée par un modèle pré-entraîné d'un réseau de neurones profond *comme KPconv* [6], [11], [19] pour extraire des représentations et *caractéristiques réels (Xr : Real Features)* . L'embedding des prototypes de classe est généré par *MoE Générateur bloc*, qui correspond à un générateur permettant de produire des *représentations générés ($X_f$ : Fake Features)* de classes non vues. Ainsi, cette approche permet de rapprocher les classes P de nuages de points 3D et leurs prototypes respectifs dans un espace latent commun, facilitant la généralisation aux classes non vues $U$. Dans la formulation du problème GZSL, pour le jeu de données de l'entraînement, nous considérons le cas du GZSL inductif, où aucune donnée des classes non vues $U$ n'est disponible pendant l'entraînement, seuls leurs prototypes de classe sont accessibles lors du test. Ainsi, l'ensemble d'entraînement est composé des triplets : $(P_i, y_i, t_{y_i})_{i \in I^S}$ où :

- $P_i \in P^S = (P_i)_{i \in I^S}$ représente un nuage de points 3D

- $y_i \in Y^S = (Y_i)_{i \in I^S}$ est l'étiquette de classes vues $S$ correspondante,
- $(t_{y_i})_{i \in I^S} \in T^S$ est le prototype de classes vues $S$ associé.

Les prototypes des classes non vues U $(t_{y_i})_{i \in I^U} \in T^U$ ne sont pas utilisés pendant l'entraînement, mais uniquement lors de la phase de test.

Donc, nous utilisons un ensemble de test pour tester le modèle sur un ensemble de nuages de points 3D $P_{test} = (P_i)_{i \in I_{test}} \in P$ étiquetés par $Y_{test} = (y_i)_{i \in I_{test}} \in Y$ où $I_{test}$ est l'indice des échantillons de test. Nous avons deux cas possibles : **(1) Si** $Y_{tes}$ contient uniquement des classes non vues **(U)**, alors nous sommes dans le cadre classique du *Zero-Shot Learning (ZSL)*. **(2) Si** $Y_{test}$ contient à la fois des *classes vues (S) et non vues (U)*, alors nous sommes dans le *cadre du Generalized Zero-Shot Learning (GZSL)*. Dans le cas de la segmentation sémantique des nuages de points 3D qui est particulièrement pertinente dans des scènes complexes et d'environnement de collaboration Humain-robot contenant plusieurs nuages de points 3D de classes différentes. Dans la pratique, les classes vues et non vues apparaissent souvent simultanément dans une scène, ce qui rend le cadre GZSL plus réaliste et mieux adapté à ce type de tâche. C'est pourquoi nous considérons la segmentation sémantique uniquement dans le cadre GZSL.

**L'architecture GZSL-MoE** repose sur trois blocs entraînés séquentiellement (voir Figures 2 et 3) :

**1)** *Feature Extraction Backbone bloc* traite les nuages de points 3D pour extraire des caractéristiques réelles ( $X_r$ : *Real Features*). Chaque objet d'un ensemble de $(P_i)_{i \in I} \in P$ est représenté par un ensemble de points 3D noté $P_i$. Pour la segmentation sémantique, $P_i$ est le nuages de points contenant le point $(P_i)_{i \in I}$ à étiqueter. Le backbone est une méthode segmentation sémantique comme KPconv [11], [19], RandLA-Net [20], [21]. Elle extrait plusieurs représentations des caractéristiques réelles ($X_r$ : *Real Features*) et une pour chaque point cloud $(P_i)_{i \in I} \in P$ pour la segmentation. Pour la phase de l'entraînement du backbone, nous entraînons d'abord le backbone des caractéristiques réelles sous supervision complète (*supervised training*) des classes vues, en l'associant à *un classificateur de mélanges d'experts (MoE classifier 1)*. Pour chaque exemple de classes vues $S : P^S = (P_i)_{i \in I^S}$ avec $I^S = \{i \in I | y_i \in S\}$ et leurs étiquettes de classe correspondantes $Y^S = (Y_i)_{i \in I^S}$ d'une classe vue, nous comparons la sortie de classificateur de mélanges d'experts (*MoE classifier 1*) à $y_1 = Y^S = (Y_i)_{i \in I^S}$ d'une classe vue via une perte d'entropie croisée (*Cross-Entropy Loss*). Cela permet d'entraîner conjointement le backbone et de classificateur de mélanges d'experts (*MoE classifier 1*). Après l'entraînement, seul le modèle pré-entrainé de Backbone est sauvegardé en tant que modèle pré-entrainé pour des caractéristiques réelles ($X_r$ : *Real Features*), tandis que le *MoE classifier 1* est ignoré et nous ne l'utilisons plus.

**2)** *MoE Feature Generator (GMoE) bloc* est le générateur *GMoE* de caractéristiques pour entrainer à générer des caractéristiques artificielles ($X_f$ : Fake Features) à partir d'un prototype de $(t_{y_i})_{i \in I^S} \in T^S$ classes vues $S$, et $(t_{y_i})_{i \in I^U} \in T^U$ classes non vues. Le rôle du générateur *GMoE* est de créer un ensemble d'entraînement composé de représentations synthétiques mais réalistes d'objets non vus. Ces caractéristiques générées permettront d'entraîner un classificateur final pour les classes non vues. Le principe du générateur *GMoE* permet de générer des caractéristiques (pour les classes non vues $U : P^U = (P_i)_{i \in I^U}$ avec $I^U = \{i \in I | y_i \in S\}$ et leurs étiquettes de classe correspondantes $Y^U = (Y_i)_{i \in I^U}$) qui doivent être similaires à celles extraites par le backbone, c'est-à-dire qu'elles doivent ressembler aux caractéristiques réelles des objets non vus, comme si nous avions eu accès à ces données. Contrairement au Backbone, qui est déterministe par entrainement supervisé, le générateur *GMoE* produit des représentations *générés ($X_f$ : Fake Features) de classes non vues selon* : $X_f = GMoE(Z_j, t_{y_i})$; *où $Z_j$ est un vecteur aléatoire, et à partir d'un prototype de $(t_{y_i})_{i \in I^S} \in T^S$ classes vues $S$, et $(t_{y_i})_{i \in I^U} \in T^U$ classes non vues.* Le générateur *GMoE* est entraîné pour produire des caractéristiques $X_f$ de manière à ce qu'elles soient similaires aux caractéristiques $X_r$ du backbone. L'entrainement repose sur des exemples de classe : $(t_{y_i})_{i \in I^S} \in T^S$ classes vues $S$, et $(t_{y_i})_{i \in I^U} \in T^U$ classes non vues, permettant ainsi d'apprendre à générer des caractéristiques réalistes à partir des prototypes de classe. Pour cet objectif, nous introduisons un modèle MoE dans le *GMoE générateur* bloc pour les avantages du MoE sur : (1) l'efficacité computationnelle [12], [16], [22], [23], où seuls quelques experts sont activés à chaque fois, ce qui réduit le coût de calcul, (2) la spécialisation où chaque expert se concentre sur une partie spécifique des données, améliorant ainsi la performance, (3) la scalabilité où le MoE est utilisé dans des modèles massifs, comme Mistral [23] et DeepSeek [14], [18] pour gérer des tâches complexes avec un grand nombre de paramètres. L'intégration de modèle MoE dans le GMoE générateur permet de mieux entraîner le générateur pour l'extraction efficace des caractéristiques $X_f$ de manière à ce qu'elles soient similaires aux caractéristiques $X_r$. Cela permet d'améliorer la capacité de généralisation du modèle génératif GZSL, ci qui va également permettre au *DMoE* discriminateur de séparer entre les *classes vues et non vues*, et va être capable de classifier des échantillons de test lors de la phase de test.

**Principe du MoE** : un modèle MoE se compose : (1) un ensemble d'experts $\{E_1, E_2, \ldots, E_n\}$. Ces experts sont des réseaux neuronaux standard de *FeedForward Neural (FFN)* spécialisés dans différentes sous-tâches ; (2) un réseau de *Gating G(x)* qui attribue un poids à chaque expert en fonction de l'entrée x. Seuls les *FFN* experts les plus pertinents sont activés; et (3) une combinaison des sorties : les prédictions des experts activés sont pondérées et combinées pour produire le résultat final. L'architecture GMoE basée sur *Mixture-of-Experts (MoE)* repose sur une couche *MoE* (Figure 3).

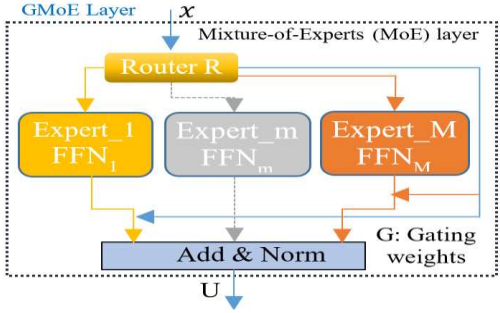

Figure 3. GMoE: Architecture basée sur Mixture-of-Experts

La structure de la couche Mélange d'Experts (MoE) utilisée ici est similaire à celle du **Mixtral Transformer** [12], [23]. Elle repose sur un réseau de **Gating G**, similaire à ceux utilisés dans **Switch Transformer** [24] **et GShard** [25] dans le module de **Routage (R)** et pour la sélection d'Experts. Le module de routage noté **R** joue le rôle du **réseau de Gating**, qui sélectionne dynamiquement les K experts (le nombre d'experts sélectionnés : 2, 4, 8, 16, etc.) les plus pertinents parmi le M experts (le nombre totale d'experts : 8, 16, 32, 64, etc.). Les experts sont représentés par des FFN (Feed-Forward Networks). Pour calculer de la sortie d'une couche MoE pour une entrée donnée x dans les couches MoE, la sortie des couches **GMoE** est obtenue par une somme pondérée des sorties des experts sélectionnés $E = \{E_1, E_2, \ldots, E_m, \ldots, E_M\}$, et la sortie est donnée par l'Équation 1 :

$$U = \sum_{m=1}^{K} G(x)_m . E_m(x) \quad (1)$$

où : K est l'ensemble des experts sélectionnés parmi les M experts, $G(x)_m$ est le poids attribué à l'expert $E_m$ par le réseau de Gating, et $E_m(x)$ est la sortie de l'expert $E_m$. Il existe plusieurs méthodes pour implémenter un réseau de Gating G(x) dans MoE, comme discuté dans de nombreux travaux. Cependant, une approche simple et efficace consiste à utiliser une couche MoE avec un gating sparse (sparsely-gated MoE layer), en prenant la fonction de Softmax sur les Top-K logits d'une couche linéaire FFN. Cette méthode est utilisée dans **Mixtral Transformer** [12], [23] et elle donnée par l'Équation 2 :

$$G(x) = Softmax\left(TopK(x, W_g)\right) \quad (2)$$

où : **TopK** sélectionne les K plus grandes valeurs des logits $l_i$ issus du produit scalaire entre l'entrée x et les poids $W_g$ de la gate. La notation de **TopK** est définie comme suit : $TopK(X_L, W_g) = (TopK(l))_i = l_i$ :

si $l_i$ est parmi les K plus grandes valeurs de $l_i \in R^M$ ; sinon $(TopK(l))_i := -\infty$. Lors de l'interprétation du modèle GMoE, l'ensemble des K experts sélectionnés parmi les M experts joue un rôle important dans l'entrainement du modèle GMoE pour générer des caractéristiques artificielles ($X_f$ : Fake Features) à partir d'un prototype de $(t_{y_i})_{i \in I^S} \in T^S$ classes vues **S**, et $(t_{y_i})_{i \in I^U} \in T^U$ classes non vues **U**. Le paramètre K (nombre d'experts sélectionnés pour chaque entrée de la séquence des nuages de points 3D) est un hyper-paramètre crucial qui contrôle à la fois la quantité de calcul utilisée pour traiter chaque entrée, l'efficacité et la sparsité du modèle.

Dans Mixtral [23], ils ont implémenté l'architecture *SwiGLU* [26] comme fonction d'expert avec M=8 experts et K=2 experts sélectionnés. Chaque expert est lui-même constitué de deux couches FFN, régulées par SwiGLU. Mais comme dans nos travaux précédents [16], [22], nous utilisons ici la fonction d'activation GELU [27] comme fonction d'expert pour réguler chaque expert, qui est constitué de deux couches FFN comme suit par l'Équation 3 :

$$E_m(x) = GELU_m(x) \quad (3)$$

*La sortie de GMoE est donnée par l'Équation 4 :*

$$U = \sum_{m=1}^{K} Softmax\left(TopK\left(\ ,W_g\right)\right)_m . GELU_m(x) \quad (4)$$

**3)** **DMoE Discriminateur (classificateur 2) bloc** pour prédire une classe $Y = (Y_i)_{i \in I}$ à partir des *caractéristiques X*. L'architecture **DMoE** est similaire à l'architecture **GMoE** basée une couche **MoE**. Le classificateur **DMoE** est entraîné à l'aide d'une perte d'entropie croisée (**Cross-Entropy Loss**) afin de prédire correctement la classe d'un point 3D à partir de ses caractéristiques générées. Dans ce cas du **Generalized Zero-Shot Learning (GZSL)**, le classificateur **DMoE** est entraîné sur un ensemble mixte contenant de deux vecteurs : (1) $X_f$ : (Fake Features) des caractéristiques artificielles générées pour les classes non vues **U** : $P^U = (P_i)_{i \in I^U}$, et (2) $X_r$(Real Features) représente les caractéristiques des classes vues ($Y^S = (Y_i)_{i \in I^S}$) extraites par le backbone (une méthode segmentation sémantique comme **KPconv** [11], [19] et **RandLA-Net** [20], [21]). La discrimination réussie pour GZSL-MoE entre deux points 3D de classes vues S et non vues U est réussie sous certaines conditions pour une classification efficace si : (1) la distribution des caractéristiques générées $X_f = GMoE(\ _j, t_{y_i})$ est similaire à celle des caractéristiques réelles $X_r$ (*Real Features*). (2) l'espace de représentation des objets 3D ou des nuages de points 3D est linéairement séparable pour chaque classe (un point 3D), ce qui permet au classificateur d'effectuer des prédictions précises donnée par la sortie $y_2$. Lors de l'inférence X, l'objectif est d'assigner une classe à un nuage de points 3D donnée en entrée du modèle en utilisant les modèles entraînés précédemment. Lors de l'inférence, étant donné un nuage de points P 3D, on utilise le backbone pré-entraîné pour extraire les caractéristiques de P. ensuite, les caractéristiques extraites sont ensuite classées par le classificateur **DMoE**. Formellement, l'inférence consiste à calculer la sortie de classificateur **DMoE** pour un nuage de points P. Le temps d'inférence et la complexité mémoire sont principalement ceux du backbone, car : le générateur **GMoE** n'est utilisé qu'à l'entrainement, pour générer des caractéristiques des classes non vues. À l'inférence, on ne fait que propager les données dans le backbone et le classificateur **DMoE**. Cette approche permet une inférence efficace, sans nécessiter de génération supplémentaire après l'entraînement. L'entraînement séquentiel de ces trois modules permet de généraliser aux classes non vues, en exploitant les prototypes de classe pour générer des représentations synthétiques exploitables par le classificateur *DMoE*.

## 3 Expériences menées GZSL-MoE

Cette section détaille l'adaptation de l'architecture GZSL-MoE appliquée aux jeux de données COVERED [8], l'entraînement du modèle 3D GZSL-MOE. Ensuite, nous présentons les résultats de la segmentation sémantique et les performances de modèle sur de scénarios de données COVERED.

### 3.1 Architecture GZSL-MoE appliquée aux jeux de données COVERED [8]

Le jeu de *données COVERED* [8] *(CollabOratiVE Robot Environment Dataset pour la Segmentation Sémantique 3D)* a été développé par le *MINDLab (Mutual Human-Robot Interaction Development Laboratory)* de l'Université des Sciences Appliquées de Zurich (ZHAW). Ce jeu de données LiDAR 3D, nommé COVERED, est fourni par MINDLab pour la segmentation sémantique 3D à l'aide de capteurs LiDAR. COVERED, tel que présenté dans [8], est un nouveau jeu de données pour l'industrie 5.0, intégrant l'usage de robots collaboratifs (Cobots) pour la Collaboration Humain-Robot (HRC) et Industrie 5.0. L'efficacité et l'utilité des robots collaboratifs (Cobots) dans un environnement de travail sécurisé permettent aux humains de compter sur ces Cobots (voir Figure 4). En effet, leur compréhension du contexte sémantique de leur environnement industriel s'améliore rapidement. Toutefois, atteindre une collaboration Humain-robot (HRC) plus sûre dans l'Industrie 5.0 nécessite des ressources de plus en plus importantes. Le développement de robots plus intelligents repose sur leur capacité à comprendre leur environnement. Or, il existe un manque de jeux de données suffisants pour la segmentation sémantique 3D, en particulier pour les interactions *Humain-Cobot*. Malgré les nombreuses recherches en cours, ce manque de données est souvent qualifié de "problème de faim de données" ("*Data Hunger*"), car il limite les performances des modèles et freine les avancées dans ce domaine. Les données COVERED sont capturées au format *PCD* (*Point Cloud Data*), également converties en format *NumPy* par capteurs LiDAR utilisés de *type Ouster OS0-128 LiDAR* pour 4 scénarios (une séquence de frames LiDAR) dans une zone de 24 m². Le référentiel contient 218 fichiers annotés (frames) au niveau point par point, disponibles en : *Format PCD (\*.pcd) et Format NumPy (\*.npy)*. Il y a 6 classes d'objets à annoter pour chaque scène dans les 4 scénarios capturés. L'indexation des classes sont (voir Figure (4-b)): *(0) label 0 : Non étiqueté, (C1) label 1 : Sol (Floor), (C2) label 2 : Mur (Wall), (C3) label 3 : Cobot (Robot collaboratif), (C4) label 4 : Humain (Human), (C5) label 5 : Véhicule AGV (Automated Guided Vehicle)*. Nous utilisation de KPConv [11], [19] pour la segmentation sémantique des nuages de points 3D. Un exemple de distribution des points 3D par classe dans un frame de données : *(C1) Sol : 10 000 points, (C2) Mur : 13 400 points, (C3) Robot : 1 800 points, (C4) Humain : 2 800 points, (C5) Véhicule AGV : 1 200 points*.

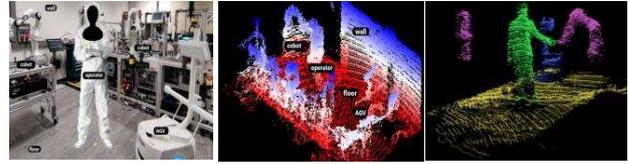

(4-a)   (4-b)   (4-c)

Figure 4. Illustration de COVERED [8] : (4-a) L'environnement industriel [8] ; (4-b) L'environnement industriel représenté sous forme de nuage de points 3D avec cinq classes d'objets : *(C1) Floor (Sol), (C2) Wall (Mur), (C3) Operator=Humain, (C4) Cobot (Robot collaboratif) et (C5) AGV = Vehicle (Véhicule à guidage automatique)*; et (4-c) Segmentation sémantique [8] des données de nuage de points 3D.

L'un des principaux défis de la collaboration Humain-robot (HRC) reste le problème de faim de données ("***Data Hunger***"). Nous introduisons l'architecture *3D GZSL-MoE* pour améliorer les limites des performances des modèles d'apprentissage de segmentation sémantique 3D des nuages de points malgré le manque de données. Pour cet objectif, nous allons donner une description détaillée de l'architecture *3D GZSL-MoE* (voir Figure 5) de ce modèle *d'apprentissage généralisé zéro-shot (GZSL) basé sur le mélange d'experts (MoE) pour la segmentation sémantique de nuages de points 3D appliqué à COVERED* comme jeu de données d'environnement de collaboration Humain-robot. Le *Backbone KPConv* traite les nuages de points 3D pour extraire des caractéristiques réelles ($X_r$: Real Features) à partir d'un ensemble de données de nuages de points 3D (PCD) noté $P_i$ pour la segmentation sémantique. Les caractéristiques réelles ($X_r$: Real Features) sont issues des classes vues S, qui incluent *Mur, Robot et Humain*. Par conséquent, pour l'entraînement *du Backbone KPConv (Kernel Point Convolution)*, nous ignorons totalement toutes les classes si elles sont non vues U, c'est-à-dire *Sol et Véhicule*. Ainsi, les données (PCD) servent d'entrée à l'extracteur de caractéristiques. Le rôle de KPConv est de classifier plusieurs points issus des données de nuages de points PCD pour la segmentation sémantique. Cependant, nous nous intéressons principalement aux caractéristiques réelles ($X_r$ : Real Features) extraites au cours de ce processus, car ils seront utilisés pour l'inférence. D'autre part, pour le générateur GMoE, les informations auxiliaires correspondent aux entrées de l'encodage utilisées *par le générateur GMoE, qui prend en entrée : un vecteur aléatoire (Ty) et des Word Embeddings issus de Word2Vec* de manière similaire à [6]. En général, l'encodage (Ty) est utilisé comme entrée, en combinaison avec le vecteur aléatoire (Z). À partir de ces deux entrées, le générateur *GMoE* produit des représentations *générés ($X_f$: Fake Features) de classes non vues **U** selon* : $X_f = GMoE(z_j, t_{y_i})$; *où $Z_j$ est un vecteur aléatoire, et à partir d'un prototype de classes vues **S (C2, C3 et C4), classes non vues U (C1 et C5)**.*

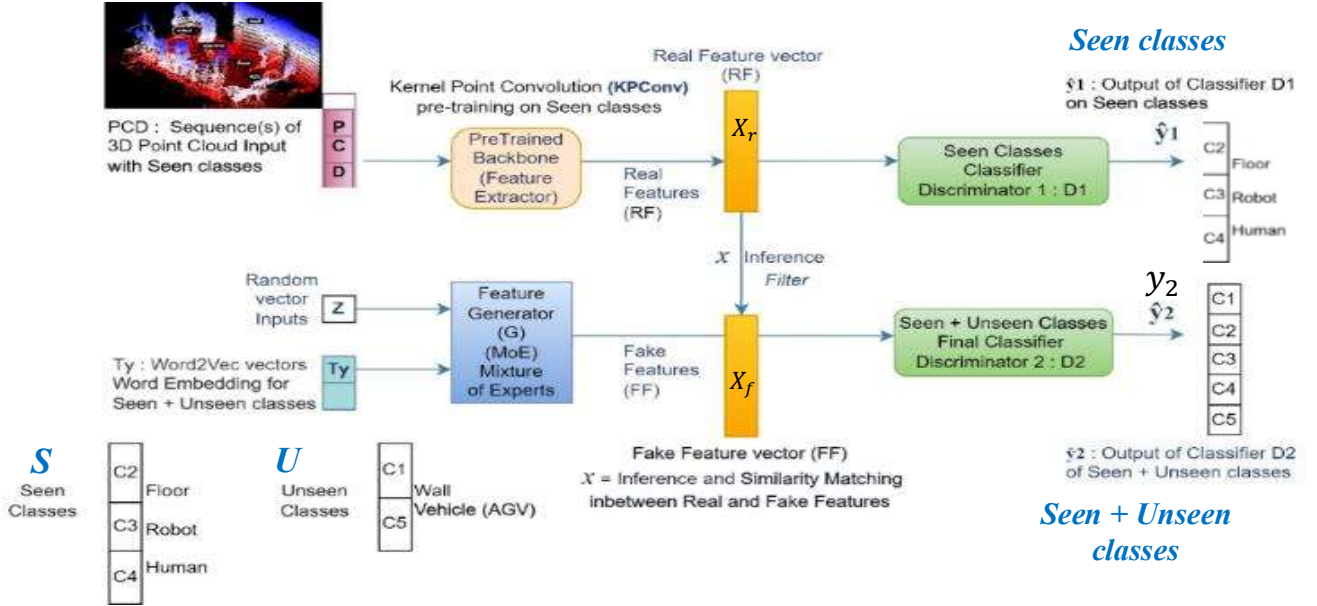

Figure 5. Description détaillée de l'architecture 3D GZSL-MoE : modèle d'apprentissage généralisé zéro-shot (GZSL) basé sur le mélange d'experts (MoE) pour la segmentation sémantique de nuages de points 3D appliqué à COVERED.

Pour être précis, Ty est composé de deux sous-ensembles de vecteurs : à partir d'un prototype de $(t_{y_i})_{i \in I^S} \in T^S$ classes vues *S*, et $(t_{y_i})_{i \in I^U} \in T^U$ classes non vues *U* qui sont tous deux générés par Word2Vec. L'inférence (X) se produit après que le Backbone (KPConv) a extrait les caractéristiques réelles et les a projetées dans le même espace latent que les caractéristiques artificielles ($X_f$ : *Fake Features*) générées par le Générateur *GMoE*. Troisièmement, nous concentrons sur le Discriminateur 2, (D2) qui utilise la fonction de **Cross-Entropy Loss** pour classifier le vecteur de caractéristiques (filtré par l'Inférence x) en sortie sous forme de labels ŷ₂=$y_2$, correspondant aux classes vues (*S*) et non vues (*U*). L'inférence (qui passe par le Discriminateur 2) est constituée de vecteurs de caractéristiques projetés dans le même espace latent. Entrainer et généraliser des classes vues (S) aux classes non vues (*U*) est impossible sans aucun exemple préalable. C'est pourquoi nous nous appuyons sur les vecteurs Ty, qui servent d'informations auxiliaires, sous forme d'un prototype unique de dimension D. *Par exemple, Ts correspond aux vecteurs (w2c) des classes vues (S), tandis que Tu représentes les vecteurs (w2c) des classes non vues (U).* Les étiquettes de ces classes ont été utilisées comme entrées dans *Word2Vec pour chacune d'elles*. L'ensemble d'entraînement est composé uniquement de classes vues (S), tandis que l'ensemble de test comprend à la fois des classes vues (S) et non vues (U). Enfin, nous pouvons comparer ces labels ŷ₂ avec les valeurs de vérité terrain (Ground Truth) et calculer des métriques d'évaluation.

### 3.2 Paramètres d'entraînement du modèle

D'abord, nous discutons des hyper paramètres d'entraînement de KPConv sur COVERED afin d'obtenir le modèle pré-entraîné KPConv avec une performance optimale pour l'extraction des caractéristiques réels de GZSL-MoE. Nous avons suivi l'entraînement sur trois scénarios (Scénario 1, Scénario 2 et Scénario 3 inclus) de COVERED et procédé à la phase de test entièrement sur le dernier scénario (Scénario 4). Les hyper paramètres étaient globalement similaires pour plusieurs configurations de taille de batch (**Batch size**) et la dimension d'entrée (Feature dimension) de **KPConv** dans ces expériences. Dans la Table 1, Nous présentons trois modèles **KPConv** pré-entraînés avec leur taille et leur précision de test. Nous pouvons observer que le premier modèle atteint le meilleur pic de précision en test ; il est aussi le plus petit en termes de taille. Néanmoins, ce premier modèle affiche des performances remarquables en comparaison, en particulier si l'on considère la petite taille de son modèle pré-entraîné.

Table 1. Performance de modèles KPConv pré-entraînés

| N° Pretrained KPConv | batch size | features dimension | num layers | Validation accuracy | max epoch | Test Accuracy | Model Size |
|---|---|---|---|---|---|---|---|
| **Model 1** | **8** | **32** | **7** | **95.31%** | **18** | **90.63%** | **200 MB** |
| Model 2 | 16 | 64 | 7 | 96.88% | 21 | 83% | 1 GB |
| Model 3 | 8 | 128 | 7 | 97.30% | 98 | 85.30% | 4 GB |

La précision en test (*Test Accuracy*) réalisée sur les 3 modèles montre que le premier modèle s'en sort très bien, et ce, avec une taille contenue de moins d'un gigaoctet (200 Mo, pour être précis). Le modèle **KPConv** le plus performant s'avère être le plus léger de tous, prenant en compte 8 lots (soit la moitié de la taille par rapport aux autres) et utilisant 32 caractéristiques en entrée pour la première dimension d'entrée, ce qui le rend globalement plus léger que les autres modèles testés jusqu'à présent. Nous avons utilisé 79 % des nuages de points du jeu de données pour l'entraînement (Scénarios 1 à 3), tandis que 21 % ont été consacrés à la validation et aux tests (Scénario 4). Le premier modèle entraîné avec **KPConv** reste le meilleur parmi tous, car ses performances surpassent les autres modèles en précision d'entraînement et en précision de validation. Ainsi, ce modèle s'impose face aux autres en obtenant les meilleures performances en précision de test. Enfin, nous avons choisi le Modèle 1 comme le meilleur modèle pré-entraîné KPConv en tant le Backbone Feature bloc pour sa taille, sa frugalité et sa complexité avec le modèle d'apprentissage GZSL. Ensuite, nous discutons des hyper paramètres d'entraînement du modèle **GZSL-MoE**. Nous avons d'abord fixé l'architecture du générateur GMoE [34] en utilisant une séquence de couches *MoE (Mixture-of-Experts layer)* avec deux configurations pour M=8 (K=2) et M=32 (K=8) experts et l'optimiseur *ADAM* pour l'entraînement du générateur avec les paramètres suivants : Taux d'apprentissage (LR) fixé à 0,0005, les valeurs de bêtas sont définies par le couple (0,92, 0,98), Pénalité de régularisation (**Weight Decay**) : 0,0001. Le modèle de génération *GMoE* possède un nombre total de paramètres entraînables de 102 millions de paramètres). Nous avons ensuite fixé l'architecture du DMoE (D2), constituée de deux couches MLP avec des couches MoE et une fonction **Softmax** pour classifier chaque lot de nuages de points. La fonction de perte utilisée pour ce DMoE est la **Cross-Entropy Loss**. Le modèle **DMoE** possède un nombre total de paramètres entraînables de 505 millions paramètres.

## 4  Résultats et Performances

Pour mesurer la performance de modèle GZSL-MOE, nous avons des classes déséquilibrées dans les ensembles de données d'entraînement, où le nombre total des deux classes non vues (*U*nseen) (Véhicule et Sol) représente 61,65 % des 262144 nuages de points des ensembles d'entraînement, tandis que les trois classes vues (*S*een) (Mur, Robots/Cobots, Humain) représentent 38,35 %. Ces classes vues présentent un déséquilibre (une classe déséquilibrée nécessite plus d'ajustements qu'une classe équilibrée), avec 120298 nuages de points associés à la classe Mur, ce qui est d'un ordre de grandeur significativement différent par rapport aux 16159 nuages de points pour la classe Robots/Cobots et aux 25137 nuages de points pour la classe Humain. Les classes non vues (*U*nseen) (Véhicule et Sol) présentent également un déséquilibre, avec 89776 nuages de points associés à la classe Sol, un volume nettement supérieur aux 10774 nuages de points de la classe Véhicule (AGV). Un ensemble de données déséquilibré peut nécessiter des stratégies supplémentaires pour un entraînement efficace, comme l'ajustement des poids des classes afin d'assurer que le modèle fonctionne de manière optimale sur de nombreuses classes déséquilibrées, telles que les classes Mur et Robots/Cobots. Dans les ensembles de validation, nous avons validé avec un total de 61,65 % (20202 nuages de points) pour les deux classes non vues (Unseen) (Véhicule et Sol) parmi les 32768 nuages de points des ensembles de validation, et 38,35 % (12566 nuages de points) pour les trois classes vues (*S*een) (Mur, Cobots et Humain). Ces classes vues se répartissent en 7865 nuages de points pour la classe Mur, 6583 nuages de points pour la classe Robots/Cobots, et 5754 nuages de points pour la classe Humain. Concernant les classes non vues (*U*nseen) (Véhicule AGV et Sol), nous avons validé avec 8652 nuages de points associés à la classe Sol et 3914 nuages de points pour la classe Véhicule AGV.
Nous évaluons le modèle GZSL-MoE avec des métriques couramment utilisées : précision globale (Acc) et précision par classe pour la classification de chaque point 3D; intersection sur union moyenne (mIoU) [8], [28] pour la segmentation sémantique. Dans le Tableau 2, nous présentons deux **modèles GZSL-MoE pré-entraînés** avec leur précision (*Accuracy*) sur l'ensemble de test. Nous observons que le premier modèle atteint le meilleur pic de précision sur l'ensemble de test.

Table 2. Performance du modèle GZSL-MoE

| N° Pretrained GZSL-MoE Model | Number of MoE layers (M) | Number of the selected experts (K) | Test Accuracy for | |
|---|---|---|---|---|
| | | | Seen Classes | Unseen classes |
| **Model 1** | **8** | **2** | **89.3 %** | **64.96 %** |
| Model 2 | 32 | 8 | 82.23 % | 56.71 % |

**Comparaison avec la méthode 3DGenZ**. Nous avons eu beaucoup de difficultés pour adapter le code de la méthode 3DGenZ pour la tester sur les données COVERED car il nous faut entrainer le modèle KPConv sur COVERED avec les mêmes hyper-paramètres imposés par la méthode 3DGenZ. La méthode de segmentation sémantique GZSL-MoE atteint un *mIoU* de 38,5 % sur les données de test (*tous S et U*) du COVERED. Les résultats détaillés sont comparés avec 3DGenZ et présentés dans le Tableau 3. La matrice de confusion dans la Figure 6 confirme que La classe non vue *AGV* est également souvent mal classifiée comme la classe vue *Cobot*.

Table 3. Performance du modèle GZSL-MoE

| GZSL méthodes | Training set | | COVERED mIoU | | |
|---|---|---|---|---|---|
| | Backbone | Classificateur | S | U | tous (U,S) |
| 3DGenZ | S | $S \cup U$ | 48.3 | 12.2 | **32.0** |
| GZSL-MoE | S | $S \cup U$ | 52.2 | 24.1 | **38.5** |

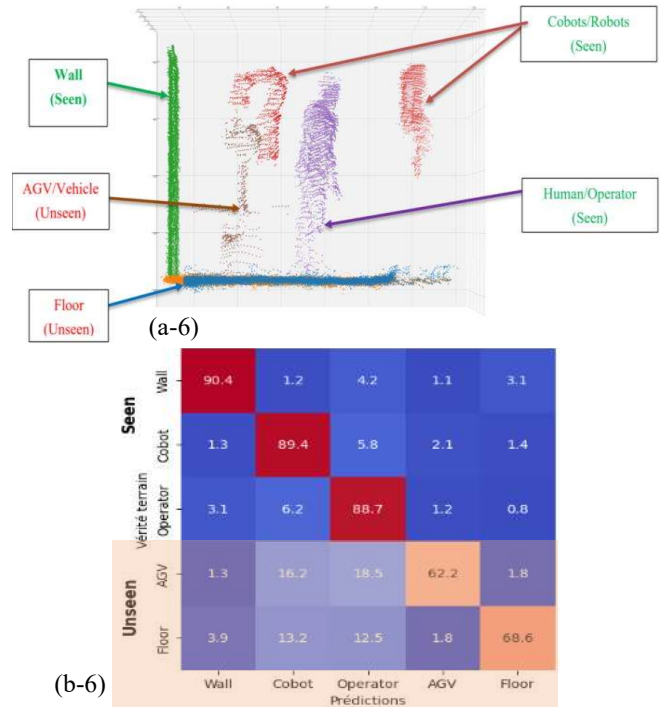

Figure 6. (a-6) Segmentation sémantique des données de nuage de points 3D COVERED par GZSL-MoE. (b-6) Matrice de confusion GZSL-MoE.

**Limitations de GZSL-MoE modèle:** Les limites de GZSL-MoE modèle proviennent du modèle pré-entraîné KPConv utilisé comme **backbone sur COVERED**. Ce modèle pré-entraîné KPConv présente certaines limites en précision de test (*Test Accuracy*) liées à ses performances sur les classes non vues pour extraire des caractéristiques car il est biaisé sur les classes vues. En utilisant du *fine-tuning*, nous avons ajusté certains hyper-paramètres (notamment le nombre de lots et la dimension des caractéristiques) afin d'atteindre les meilleures performances avec le *modèle 1 KPConv* (Table 1) pour améliorer ses performances sur les classes non vues. Cependant, nous n'avons pas testé toutes les variantes possibles de **KPConv sur COVERED**, ce qui signifie que si nous pré-entraînons le backbone avec d'autres hyperparamètres, les performances pourraient être améliorées ses performances pour extraire des caractéristiques des classes non vues.

**La seconde limitation de** GZSL-MoE provient du générateur basé sur le MoE. Nous avons effectué un fine-tuning des hyperparamètres du MoE [12], [22] avec M=8 experts par couche et K=2 experts sélectionnés par entrée. Ces paramètres ayant été ajustés spécifiquement pour le jeu de données COVERED, un réentraînement et un réglage des hyperparamètres seraient nécessaires pour une adaptation à un autre jeu de données. Les limites du modèle (même pour le meilleur modèle entraîné sur COVERED) montrent qu'un entraînement sur un autre jeu de données, comme SemanticKITTI [29] et Stanford (S3DIS) [30], pourrait donner des résultats différents.

## 5 Conclusion & Perspectives

Pour conclure, nous avons proposé le modèle d'apprentissage généralisé zéro-shot basé sur le mélange d'experts (GZSL-MoE). Ce modèle intègre des couches de mélange d'experts (MoE). L'intégration du mélange d'experts dans les composants Générateur et Discriminateur du modèle GZSL permet d'améliorer la segmentation sémantique de nuages de points 3D, appliquée au jeu de données pour des environnements de collaboration Humain-robot. Le nouveau modèle GZSL-MoE offre une solution prometteuse pour comprendre des environnements 3D complexes. Le modèle génératif GZSL-MoE proposé dans notre étude est capable d'étiqueter des classes non vues lors du test, bien qu'il ait été principalement entraîné sur des classes vues du jeu de données COVERED. Dans le cas de GZSL-MoE, comme les résultats peuvent être biaisés en faveur des classes vues, nous allons utiliser une métrique commune qui consiste à rapporter la moyenne harmonique (HM) des mesures pour les classes vues et non vues (qu'il s'agisse de *Acc ou de mIoU*). Dans nos futurs travaux, nous allons entrainer et tester le modèle GZSL-MoE avec un Backbone RandLa-Net.

## 6 Références